\documentclass[journal, letterpaper]{IEEEtran}

\usepackage[T1]{fontenc}
\usepackage{textcomp}
\usepackage{graphicx}
\usepackage{url}        
\usepackage{amsmath}    
\usepackage{authblk}

\usepackage{textgreek}	
\usepackage{listings}
\usepackage{longtable}
\usepackage{array}
\usepackage{hyperref}
\usepackage[
    style=numeric,
    sorting=none,
    defernumbers=false,
    citestyle=numeric,
    bibstyle=numeric
]{biblatex}
\DeclareFieldFormat{doi}{}
\DeclareFieldFormat{url}{}
\usepackage{tikz}
\usepackage{textcomp}
\usepackage{hyperref}
\usepackage{lipsum}
\newcommand\copyrighttext{%
  \footnotesize \textcopyright 2022 IEEE. Personal use of this material is permitted.
  Permission from IEEE must be obtained for all other uses, in any current or future 
  media, including reprinting/republishing this material for advertising or promotional 
  purposes, creating new collective works, for resale or redistribution to servers or 
  lists, or reuse of any copyrighted component of this work in other works. 
  DOI: \href{https://doi.org/10.1109/ACCESS.2022.3187002}{10.1109/ACCESS.2022.3187002}}
\newcommand\copyrightnotice{%
\begin{tikzpicture}[remember picture,overlay]
	\node[anchor=south,yshift=2pt] at (current page.south) {\fbox{\parbox{\dimexpr\textwidth-\fboxsep-\fboxrule\relax}{\copyrighttext}}};
	\end{tikzpicture}%
	}
\title{PTEENet: Post-Trained Early-Exit Neural Networks Augmentation for Inference Cost Optimization}
\author{Assaf Lahiany\textsuperscript{1}, Yehudit Aperstein\textsuperscript{1}}

\vspace{20pt}  
\affil{\textsuperscript{1}Intelligent Systems Department\\Afeka Academic College of Engineering, Tel Aviv-Yafo, Israel}

\begin{document}

\maketitle
\copyrightnotice

\begin{abstract}
	For many practical applications, a high computational cost of inference over deep network architectures might be unacceptable. A small degradation in the overall inference accuracy might be a reasonable price to pay for a significant reduction in the required computational resources. In this work, we describe a method for introducing "shortcuts" into the DNN feedforward inference process by skipping costly feedforward computations whenever possible. The proposed method is based on the previously described BranchyNet \cite{teerapittayanon2016branchynet} and the EEnet \cite{demir2019early} architectures that jointly train the main network and early exit branches. We extend those methods by attaching branches to pre-trained models and, thus, eliminating the need to alter the original weights of the network. We also suggest a new branch architecture based on convolutional building blocks to allow enough training capacity when applied on large DNNs. The proposed architecture includes confidence heads that are used for predicting the confidence level in the corresponding early exits. By defining adjusted thresholds on these confidence extensions, we can control in real-time the amount of data exiting from each branch and the overall tradeoff between speed and accuracy of our model. In our experiments, we evaluate our method using image datasets (SVHN and CIFAR10) and several DNN architectures (ResNet, DenseNet, VGG) with varied depth. Our results demonstrate that the proposed method enables us to reduce the average inference computational cost and further controlling the tradeoff between the model accuracy and the computation cost.
\end{abstract}

\section{Introduction}	

Deep Neural Networks (DNN) models are frequently used for solving difficult machine learning problems in various domains including computer vision \cite{krizhevsky2012imagenet}, natural language processing \cite{mikolov2010recurrent} and sensory data processing \cite{yao2017deepsense}. Provided enough training data is available, deeper, and more complex neural architectures frequently lead to better performance \cite{simonyan2014very}. For example, for a visual object classification problem, a deeper architecture (e.g., AlexNet \cite{krizhevsky2012imagenet}) almost always leads to better performance. However, processing inputs using deeper and more complex neural networks requires a significant amount of computing power and may increase the overall processing latency \cite{he2015convolutional}, \cite{guo2018cloud}. These processing power requirements may prevent the deployment of deep learning applications at edge devices (e.g., smartphones or sensors) that have severe limitations in processing power or battery life \cite{zhang2019deep}, \cite{ota2017deep}.

Distributed processing approach allows one to alleviate computing power limitations by employing multiple computing nodes for performing necessary computations \cite{dean2012large}. For instance, the DNN computation load can be partitioned between edge devices (e.g., smartphones) and more powerful server nodes. In this case, the edge devices might perform necessary computations for extracting features from the raw input (e.g., video). Then, the extracted features can be sent to a server where the rest of the computations are performed. In the case of DNN, an edge device might compute a few lower layers of the network, while transmitting the output of an intermediate layer to the server for the rest of the processing \cite{mao2017modnn}. The distributed computing might employ more than two types of computing nodes besides the server and edge device nodes. For instance, fog computing \cite{teerapittayanon2017distributed} architectures define a hierarchy of computing nodes according to their computing power and their role within the platform’s network topology (e.g., gateways). Unfortunately, the distributed processing scheme might result in a significant network traffic due to data transmissions among computing nodes. Depending on the depth of a neural network and the partition of the network among computing nodes, the intermediate layers might produce tens of thousands of real-valued values that needed to be communicated between computing nodes \cite{ota2017deep}.

In this work we propose an optimization scheme that extends the work of BranchyNet \cite{teerapittayanon2016branchynet}. The BranchyNet concept suggests augmenting the original network with small intermediate decision networks attached to selected hidden layers within the DNN. These small networks (called Branches) are trained to infer network outputs (e.g., classification labels) on “easy” input cases. The branch networks are trained jointly to make decisions solely based on the values produced by intermediate layers of the main DNN. When such inference is possible, the inference through the rest of the DNN layers is interrupted, saving precious resources including network and computing capacities.

Our approach extends the original BranchyNet concept for supporting two practical requirements. We assume that the original training data might not be available during the optimization and deployment time. Moreover, the parameters of the training procedure for the main network might not be known, or training process might not be easily reproducible. We also assume that resource limitation and accuracy requirements are application dependent. Therefore, the augment model should have a “knob” providing explicit and predictable control over the tradeoff between accuracy and computational costs.

Motivated by these assumptions, we attach branches to an already trained network to address the cases where no complete copy of original training data is available during the optimization of the network performance. The network is augmented by branches that contained two “heads”: classification head and decision head. The classification head is trained to mimic the output of the original network and the decision head estimates the reliability or certainty of the match between the outputs of the branch and the original network. Depending on the threshold, the combine model continues to evaluate higher layers of the main network if certainty is too low. We evaluate the effectiveness of our approach on a standard benchmark: SVHN \cite{goodfellow2013multi} and CIFAR10 \cite{krizhevsky2014cifar}.

\section{Prior Work}

There are several strategies for neural network optimization that have been explored in the past. The existing approaches can be categorized into two categories: network complexity reduction \cite{cheng2017survey} and distributed computing \cite{teerapittayanon2017distributed}. In this section, we cover common methods in both categories. It should be noted that we focus on techniques for runtime performance optimization during inference as compared to runtime performance optimization of training procedures. The optimization of DNN training time is also frequently discussed in this context of distributed neural networks \cite{cheng2017survey}. Some of the common approaches for reducing the computation and communication requirement for DNN inference are based on weight quantization. Weight quantization methods enable reduction of the number of bits required for storing network weights. Reducing the number of bits results in simpler and faster computations \cite{gong2014compressing}. There are more advanced approaches for quantization that are based on weights clustering. In these approaches, a network weight is approximated by the center of its closest cluster and, thus, can be encoded by a smaller number of bits. 

Network complexity reduction can also be achieved using network pruning \cite{han2015deep} and connection sharing \cite{srinivas2015data}. Network pruning techniques reduce the network complexity by dropping some less important neurons and connections. There are different methods for selecting the most optimal pruning strategies \cite{cheng2017survey}. For example, one method suggests dropping all connections from a fully connected layer whose weights are lower than a predefined threshold \cite{han2015learning}. This technique basically converts a dense layer to a sparse layer and reduces the storage and computation requirements by an order of magnitude.

Connection sharing methods allow to reduce the complexity by sharing parameters between multiple neurons \cite{srinivas2015data}. For example, in CNN models the main assumption is that one filter which is useful for computation at some specific data range in one of the model layers, can also be useful for computation at a different data range in the same layer \cite{krizhevsky2012imagenet}. Whenever a deep learning network is deployed in a distributed environment, efficient partitions of the DNN between nodes and different communications schemes might lead to reduced communication network and processing load. For instance, \cite{mao2017modnn} suggests an efficient way for mapping sections of DNN onto a distributed computing hierarchy. In \cite{lo2017dynamic}, authors suggest to deploy shallow networks to edge devices for performing a “gating function”. The output of these shallow (auxiliary) networks is used for deciding if an input has to be transferred to the stronger backbone servers for inferring using more deep and complex neural networks.

The original approach based on early exit augmentation is based on the BranchyNet ideas \cite{teerapittayanon2016branchynet}. BranchyNet concept suggests to augment the main deep neural network with additional side branch classifiers attached to the selected intermediate layers. The augmented network has a single-entry point and multiple exit points. The decision of the augmented network can be produced at any exit point. The output of branch networks is used for reaching and early decision and stopping all further processing. If a side branch classifier indicates a certain degree of confidence, all further processing is stopped and the decision is made based solely on the output of the branch classifier. Another early exit approach is introduced in \cite{demir2019early} where a confidence head is added in parallel to the classifier head. The confidence sigmoid output is trained to produce confidence level output which is used at inference time as the early exit decision making mechanism.

Few other Early-Exit architectures and design implementations were introduced in the field of network cost optimization. \cite{ju2021learning} presents a Learning Early Exit (LEE) scheme with an online algorithm that chooses the exit point of a DNN by performing history exploration using a reward formulation. For hardware aware approaches, \cite{laskaridis2020hapi} defines HAPI framework which uses hardware-aware design of early-exit CNNs as a mathematical optimization problem. This approach generates progressive inference networks  customized and tailored for the specific target deployment platform. Furthermore, \cite{li2021developing} presents a neural network designed for miniature edge devices, which allow distributed implementation (on both flash memory and on chip SRAM memory) of a small 2 exit network. Other early exit methodologies were introduced in \cite{bonato2021class}, \cite{gormez2021class}. \cite{bonato2021class} introduces an early-exit opportunities on a reference model targeting a specific class which leads to improved average classification rate for the specific class, maintaining the original model accuracy. \cite{gormez2021class} presents an early exit mechanism based on class means. It obtains the means by taking the mean of layer output of each class at every layer of the model. During inference, output of a layer is compared with the corresponding class mean to stop execution.

\section{Our Contribution}

We extend the original BranchyNet model by introducing logic basic-blocks which in the case of ResNet architecture several critical enhancements:
\begin{itemize}
	\item We define a new early exit branch architecture which includes both classification output and confidence head for early termination decision making. Those branches are attached to the pre-trained backbone network to form a PTEEnet (Post Trained Early Exit network).
	\item Branch placement is done using suitable distribution methods to allow computational cost optimization, while considering architectural constrains of the original network.
	\item We train only the attached branches using a loss function that combines both the accumulated prediction loss and the weighted computational cost associated with the consumed computations resources. The loss function is designed to express the certainty of the branch output while considering the cost of proceeding to deeper layers. 
	\item For branch training and accuracy evaluation, we use the main pre-trained network output as a label generator to allow the use of unlabeled dataset. This allows for more practical scenarios where we don’t have the original dataset used for the original network training. 
\end{itemize}

\section{PTEE Methodology}

Our PTEE (Post Trained Early exit) model can be casted on many deep neural network architectures with various branch distribution suited for each architecture. We first describe the global PTEEnet network architecture (which is based on previous EEnet work \cite{demir2019early}) and then expand it to define the new methodology.

\subsection{EE-Blocks Distribution}

The number of the early exit branches and their placement are important factor in the model architecture. As introduced in \cite{demir2019early}, many distribution methods are suitable. Pareto method is based on the 80/20 principle where 80\% of the samples are classified by 20\% of the total computational cost and therefore define a 0.2 ratio between the added computational cost and the total cost calculate from each previous branch. In the same manner, Fine and Golden distribution methods define ratios to be at least 0.05 and 0.618 respectively depending on the internal design of the network. Linear distribution method defines a fixed computational cost gap. To reduce freedom degree to our exploration we use the Fine distribution method.

\begin{figure}[ht!]
	    \begin{center}
		\includegraphics[scale=0.5]{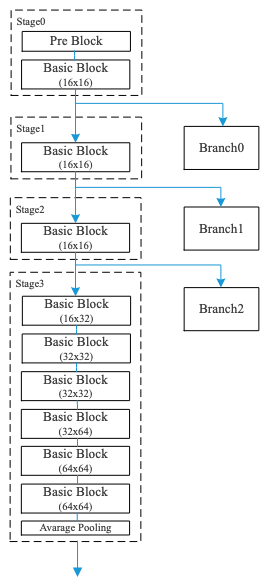}
    	\caption{ResNet20 model attached by 3 early exit branches. Each branch allows for early termination of samples propagation by applying confidence head threshold.}
    	\label{fig:blockDistribution}
    	\end{center}
	\end{figure}

Figure \ref{fig:blockDistribution} shows 3 branches distributed on ResNet20 main network, using Fine distribution method. Choosing the optimal number of branches depends mainly on main network size, distribution method and dataset characteristics. We follow the computational cost levels defined by the distribution method to place the branches along the main network. The original network is organized in stages, defined by the placement of the attached branches. These are only semantic structure to help locate branches and define network segments. Each branch and its leading stages are grouped into segments which are used for further complexity analysis. For simplicity, branches are attached only between logic basic-blocks which in the case of ResNet architecture consist of a single residual block.

\subsection{Cumulative Predictions and Computational Cost}

We begin our definition of our loss elements by using the cross-entropy function as the classification loss for each exit head:
\begin{equation}
\begin{aligned}
\label{eq:loss}
\mathcal{L}_{MC} = CE\left( y,\widehat{Y} \right) = - \sum_{n = 1}^{K}{y_{n} \cdot \log{\left( {\widehat{Y}}_{n} \right),}}
\end{aligned}
\end{equation}
where \emph{K} is the number of classes in our dataset (\emph{K}=10 for CIFAR10). The final classification output vector, \(\widehat{Y}\), can be obtained from a single equation which consists of the outputs of all exit blocks.
\begin{multline}
\label{eq:classification}
\widehat{Y} = I_{(h_{0} \geq T)} \cdot {\widehat{y}}_{0} + I_{\left( h_{0} < T \right)} \cdot \\
\left\{ \ldots I_{\left( h_{N - 1} \geq T \right)} \cdot {\widehat{y}}_{N - 1} + I_{\left( h_{N - 1} < T \right)} \cdot {\widehat{y}}_{N} \right\}\ldots
\end{multline}

where \(\widehat{Y}\) denotes the final classification output vector and \emph{N} defines the number of early-exit blocks. \({\widehat{y}}_{i}\) and \(h_{i}\) are the classification output vector and confidence score of the \(n_{th}\) early-exit block respectively. \({\widehat{y}}_{N}\) is the predicted output vector of the pre-train network classifier (early exit branches are indexed from 0 to \emph{N}-1 where \emph{N} is the index of the main backbone network). In addition, \(I_{\{ h_{0} \geq T\}}\) denotes the binary indicator function mapping a confidence score onto 0 or 1 which results in continuation or an exit decision, respectively. However, to perform backpropagation during training, we need to define a differentiable version of equation (\ref{eq:classification}) by approximating the binary indicator function. To do so, we use the continuous sigmoid function, where \(h_{n}\) denotes the confidence score of the \(n_{th}\) early-exit block. Finally, we can derive the cumulative prediction \({\widehat{Y}}_{n}\), as defined in (\ref{eq:cumulative}), where each \({\widehat{Y}}_{n},\ (n = 0\ldots.N - 1)\) defines the classification output vector derived by all proceeding exit blocks from n to \emph{N}-1. The last \({\widehat{Y}}_{n + 1}\) (\(n = N - 1\)) is \({\widehat{Y}}_{N}\), the output of the main pre-trained classifier which has no corresponding confidence head.
\begin{equation}
\begin{aligned}
\label{eq:cumulative}
{\widehat{Y}}_{n} = h_{n} \cdot {\widehat{y}}_{n} + \left( 1 - h_{n} \right) \cdot {\widehat{Y}}_{n + 1};\ \ \ \ \ \ n = 0\ldots.N - 1
\end{aligned}
\end{equation}

As a contrasting force, to maintain accuracy-cost balance for the final loss function, we define the cumulative computational cost. It should encourage the model to classify easy examples early. The computational cost of the \(n_{th}\) exit block, \(c_{n}\), is calculated by the number of floating-point operations (FLOPs) from the base of the model to the classification head of the corresponding exit block. Since FLOPs values are too large to be included in the loss function (when comparing to cross entropy loss values), they are normalized by the total number of FLOPs of the original backbone model. The general cost, \(C_{n}\), can be derived from the same cumulative approach:
\begin{equation}
\begin{aligned}
\label{eq:cost}
C_{n} = h_{n} \cdot c_{n} + \left( 1 - h_{n} \right) \cdot c_{n + 1};\ \ \ \ \ \ \ n = 0\ldots.N - 1
\end{aligned}
\end{equation}
For the total early-exit loss, we use definitions in (\ref{eq:cumulative}) and (\ref{eq:cost}) to define the total weighted cumulative loss:
\begin{equation}
\begin{aligned}
\label{eq:loss_full}
\mathcal{L} = \sum_{n = 0}^{N - 1}\mathcal{L}_{MC}^{(n)} + \lambda\mathcal{L}_{Cost}^{(n)} = \sum_{n = 0}^{N - 1}{CE\left( y,{\widehat{Y}}_{n} \right)} + \lambda C_{n}
\end{aligned}
\end{equation}

where \(\lambda \geq 0\) penalty weight. The construction of (\ref{eq:loss_full}) is motivated by the fact that in the non-cumulative weighted approach, high confidence of an early-exit block, may disable the prediction contributions of the following exit blocks to the cumulative prediction \({\widehat{Y}}_{n}\). Consequently, the following exit blocks are not trained properly without enough back-propagation signals. We need to be sure that the predictions coming from all exit blocks are trained fairly. We expect that during the training process, the confidence scores \(h_{n}\) will gradually tend to be higher value at the deeper exit points. However, the cost penalty \(\mathcal{L}_{Cost}\) and its weight \(\lambda\), forces the model to terminate early since the shallower early-exit blocks have less computational cost.

\subsection{Training}

As a preliminary stage to training our branches, we freeze the pre-trained weights of the backbone network and disable their gradients calculation. To eliminate our dependency on data labels, we use the main network outputs as the ground truth values (replacing the original labeled data). This is done by executing forward pass of each batch of training data and use the main network classifier predictions to determine the corresponding labels. These labels are used both for training and validation. The loss is back-propagated only till the stitching point of each branch and the weights are updated.

\subsection{Inference}

During inference we apply a stop criteria in the form of a dedicated confidence threshold level \emph{T} on each of the confidence heads in the early exit branches. This will be used for the early propagation termination. If the confidence score of an early-exit block is above the threshold, the classification output of the current stage will be the final prediction. Each input is classified based on their individual confidence scores predicted by the early-exit blocks. Thus, one input can be classified and terminated early while others continue to propagate through the model. The threshold-confidence mechanism can serve as “knob”, updated in real time to ensure desired accuracy-computational cost trade-off. The impact of threshold levels on model performance, should be examined on the final model a-priory to produce desired known result. The inference procedure of the early exit model is given in the following pseudo-code:
\begin{table}[!hbt]

	\begin{center}
	\renewcommand{\arraystretch}{1.2}
	\caption{PTEEnet Fast Inference Procedure.}
	\label{tab:algorithm}

	\begin{tabular}{|c|m{4cm}|}

		\hline		
		\emph{1} & \(i \leftarrow 0\) \\
		\emph{2} & \emph{\textbf{While}} \(i < N\ \)\emph{do} \\
		\emph{3} & \(\ \ \ \ \ \ \ \ x \leftarrow BasicBlocks_{i}(x)\) \\
		\emph{4} &
		\(\ \ \ \ \ \ \ h_{i},{\widehat{y}}_{i} \leftarrow {EEBlock}_{i}(x)\) \\
		\emph{5} & \(\ \ \ \ \ \ \ \mathbf{If}\ h_{i} \geq T\ then\) \\
		\emph{6} & \(\ \ \ \ \ \ \ \ \ \ \ \ \ \emph{\textbf{Return}}\ {\widehat{y}}_{i}\) \\
		\emph{7} & \(\ \ \ \ \ \ \ \emph{\textbf{End if}}\) \\
		\emph{8} & \(\ \ \ \ \ \ \ i \leftarrow i + 1\) \\
		\emph{9} & \emph{\textbf{End While}} \\
		\emph{10} & \(x \leftarrow {BasicBlocks}_{i}(x)\) \\
		\emph{11} & \(\widehat{y} \leftarrow ExitBlock(x)\) \\
		\emph{12} & \(\mathbf{Return}\ \widehat{y}\) \\
		\hline
	\end{tabular}
	\end{center}
\end{table}

where \(EEBlock_{i}\) represents the \(i_{th}\) early-exit (EE) block of the model and \(BasicBlocks_{i}\) denotes the sequence of intermediate blocks between \({(i - 1)}^{th} EEBlock\) and \(i^{th} EEBlock\). Obviously, \(BasicBlocks_{0}\) is the initial basic block of the model before entering any EE-block. \emph{N} denotes the total number of early-exit blocks. \(h_{i}\) and \({\widehat{y}}_{i}\) shows the confidence score and classification output vector of the \(i_{th}\) EE-block and \emph{T} is the confidence threshold used as a stop criteria during and control the branch throughput in terms of samples propagation from each branch.

\subsection{Branch Architecture}

\begin{figure}[ht!]
	\begin{center}
	\includegraphics[scale=.5]{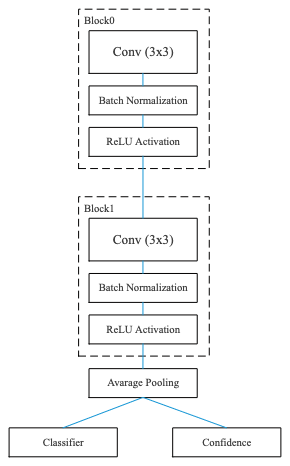}
	\caption{\emph{ConvX} branch architecture. Using both \emph{block0} and \emph{block1} defines \emph{Conv2} branch architecture. Confidence head output is used for early termination decision making by applying threshold \emph{T} on its output value. Higher that threshold \emph{T} values terminate sample propagation and the use of the attached classifier prediction.}
	\label{fig:branchArchitecture}
	\end{center}
\end{figure}

Using shallow capacity branch architecture, as in \cite{demir2019early}, is only allowed while using small size backbone networks. To account for deeper network architectures, typically ResNet20 and on, we need higher learning capacity branches. Exploring different branch architectures yield \emph{ConvX}, as presented in Figure \ref{fig:branchArchitecture}, which consist of sequential convolutional blocks attached to classifier and confidence heads by an average pooling layer. Each block consists of a convolutional layer using a 3x3 kernel followed by a batch normalization layer and a ReLU activation layer. The confidence head uses a sigmoid output and is trained to yield high values (towards 1) reflecting high confidence levels in the corresponding classifier prediction. Appling threshold value \emph{T} to the confidence head output is used during validation/test accuracy calculation, and act as a “knob” during real-time inference of the PTEEnet model, controlling accuracy-cost tradeoff.

\subsection{Complexity}

To measure the inference computational complexity of any network segment in our model we use FLOPs (floating point operations) as a common measurement unit used also in the original ResNet paper {[}32{]}. Each FLOP is defined by a pair of multiple-accumulate (MAC) operations. Those MAC operations are sometimes referred to as FMA (fused multiple accumulate) when using more modern hardware architectures. An exit network segment includes all stages (see Figure \ref{fig:blockDistribution}) preceding the branch attach point and the branch itself. To calculate the number of FLOPs in each exit segment we perform a forward pass and accumulate the total number of FLOPs from all segment layers. For example, the number of FLOPs in a 2D convolution layer with single stride and padding, square kernel of size \emph{k}, \(C_{i}\) input channels, \(C_{o}\) output channels and (\emph{H, W}) as the size of the input feature map, is \(C_{i} \times k^{2} \times C_{o} \times H \times W\). Finally, we define the branch relative cost as the ratio between the number of exit segment FLOPs (from input to branch exit) and the number of FLOPs in the main backbone network. Table \ref{tab:complexity} presents the total FLOPs and relative cost of each exit segment using VGG19, DenseNet121, and ResNet110 models. For each PTEEnet model architecture we used \emph{Fine} distribution method, \emph{Conv2} branch architecture and 32x32 input size (as in CIFAR10 and SVHN datasets). Each branch could be attached only between consecutive basic blocks of each network architecture.

\begin{table}[!hbt]

	\begin{center}
	\renewcommand{\arraystretch}{1.2}
	\caption{Complexity of PTEEnets including VGGEEnet19, DenseEEnet121 and ResEEnet110 using Conv2 branch architecture.}
	\label{tab:complexity}

	\begin{tabular}{|c|c|c|c|}

		\hline
		\hline
		Network & Exit & MFLOPs & Relative Cost \\
		
		\hline
		VGGEEnet19 & 0 & 59.10 & 0.15 \\
		& 1 & 115.96 & 0.29 \\
		& 2 & 248.33 & 0.62 \\
		& main & 400.00 & 1.00 \\
		DenseEEnet121 & 0 & 218.86 & 0.24 \\
		& 1 & 402.00 & 0.45 \\
		& 2 & 627.15 & 0.70 \\
		& main & 898.06 & 1.00 \\
		ResEEnet110 & 0 & 19.78 & 0.08 \\
		& 1 & 34.23 & 0.13 \\
		& 2 & 43.86 & 0.17 \\
		& 3 & 53.49 & 0.21 \\
		& 4 & 63.13 & 0.25 \\
		& 5 & 72.76 & 0.28 \\
		& 6 & 82.40 & 0.32 \\
		& 7 & 92.03 & 0.36 \\
		& 8 & 100.48 & 0.39 \\
		& 9 & 110.01 & 0.43 \\
		& main & 256.32 & 1.00 \\
		\hline
	\end{tabular}
	\end{center}
\end{table}

The main exit branch in each model corresponds to the main backbone network exit and yields relative cost of 1. 

\section{Experiments and Results}

To evaluate our PTEEnet methodology we use 3 state of the art vanilla backbone network architectures -- ResNet, VGG and DenseNet to create their corresponding PTEEnet variants: ResEEnet, VGGEEnet and DenseEEnet. We evaluate those PTEEnet using CIFAR10 and SVHN datasets. For the ResEENet architecture we use pre-trained ResNet20, ResNet32, ResNet110 backbones, and attach 3,5,10 branches respectively. This results in ResEEnet20/3, ResEEnet32/5, ResEEnet110/10 models. For VGG and DenseNet architectures, we use a VGG19 and DenseNet121 backbones, each attached with 3 branches to construct a VGGEEnet19/3 and DenseEEnet121/3. For each model we ``freeze'' the pre-trained weights, and train only the branches. We define cost penalty \(\lambda\) range from 0.2 to 2.3. As described in the previous section, the cost penalty \(\lambda\), defines the amount of penalty we induce on high relative cost branches during training stage. Increasing \(\lambda\) leads to a decrease in the number of calculations and as a result a decrease in accuracy.

\begin{figure}[ht!]
	\begin{center}
	\includegraphics[scale=0.5]{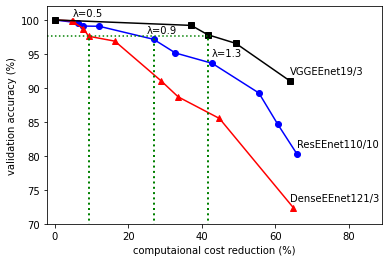}
	\caption{Computational cost reduction and validation accuracy pairs generated from increasing levels of \(\mathbf{\lambda}\), each used to train ResEEnet110, VGGEEnet19 and DenseEEnet121 models with various number of branches attached. \(\mathbf{\lambda}\) values correspond to the blue curve of ResEEnet110.}
	\label{fig:computationalCost}
	\end{center}
\end{figure}

Figure \ref{fig:computationalCost} presents accuracy and computational costs for different values of penalty parameter \(\lambda\) in three PTEEnet models. Validation accuracy and cost reduction pairs calculated using the stop inference criteria with fixed confidence level threshold of \(T_{0} = 0.5\).

Blue curve in Figure \ref{fig:computationalCost} presents accuracy and computational cost reduction behaviour of ResEEnet110/10 on CIFAR10 dataset. The model trained with \(\lambda = 0.5\) produced \textasciitilde8\% computational cost reduction while maintaining 99\% validation accuracy; \(\lambda = 0.9\) yield \textasciitilde27\% cost reduction and maintained accuracy of 97.15\%. Increasing \(\lambda\) to 1.3 produced 42\% computational cost reduction and accuracy level of 93.6\%. The black and red curves correspond to the results of VGGEEnet19/3 and DenseEENet121/3 respectively on the same data. For fixed accuracy level of 97\% computational cost reduction (horizontal dotted line) levels for each PTEENet: ResEEnet110/10 - 27\%, VGGEEnet19/3- 42\%, and DenseEEnet121/3 - 9\%. The major reduction of computational cost using CIFAR10 on VGGEEnet19/3 can indicate that the original capacity size of the VGG19 is more than enough to handle the samples classification difficulty.

Setting a parameter value \(\lambda\) can be used for selecting the adequate accuracy-cost ratio suitable for deploying a system in real-world scenarios. For instance, if the goal is to maximize ration between the computational cost reduction and the accuracy reduction, for ResEEnet110/10 on CIFAR10 then \(\lambda = 1.3\) is an optimal value and it results in 6\% accuracy drop while achieving almost 42\% computational cost reduction.

As discussed in section IV E, during inference the decision about a sample class is made based on confidence threshold level \emph{T} on each of the confidence heads in the early exit branches.

\begin{figure}[ht!]
	\begin{center}
	\includegraphics[scale=0.5]{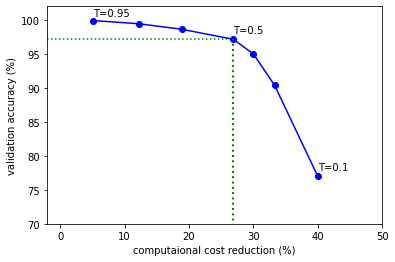}
	\caption{Computational cost reduction and validation accuracy pairs generated from varied levels of confidence threshold \emph{T}, for ResEEnet110 model with 10 branches attached. \emph{T}=0.5 maintain accuracy of 97.15\% while gaining computational cost reduction of 27\%}
	\label{fig:computationalCost2}
	\end{center}
\end{figure}

Since \emph{T} can be externally controlled during inference it can act as a ``knob'' to control the real-time accuracy-cost tradeoff. In this work we set the same threshold value to all branches, although further optimization of each branch threshold can be done. Using the previous ResEEnet110/10, trained with \(\lambda = 0.9\), we evaluate the accuracy-cost pairs for each value of \emph{T} in the range of 0.1 to 0.95. Figure \ref{fig:computationalCost2} presents the accuracy-cost reduction pairs using different levels of \emph{T}. The interpolated curve has a ``knee'' like shape with slow decrease in accuracy for \(T \geq 0.5\) and significant rapid decrease in accuracy for \(T < 0.5\). The confidence threshold \(T^{*} = 0.5\) is an optimal level for maintaining high accuracy with large reduction in computational cost.

The final optimal confidence level threshold selection depends on the application requirements in terms of the amount of accuracy decrease allowed and computational resources at hand.

Table \ref{tab:costReduction} presents a performance comparison of various PTEEnet models based on maximum 3\% drop tolerance in validation accuracy. For each model we set a \(\lambda\) value that produced the maximum reduction in computational cost and while using fixed confidence threshold of \emph{T}=0.5 for validation accuracy and cost calculation. ResEEnet110/10 produced 25\% cost reduction while ResEEnet20/3 and ResEEnet32/5 had only 15\% reduction in cost. It seems that using high number of branches on high-capacity backbones allow for more ``accurate'' exit options in terms of minimal loss and higher confidence for the propagated samples thus reducing computational load. Different datasets with different complexity levels yields different cost reduction levels, as can be seen using VGGEEnet19 and DenseEEnet121 on CIFAR10 and SVHN. This is probably due to the different distribution of classification difficulty of the samples in each dataset. VGGEEnet19 on CIFAR10 produce 42\% cost reduction compared to SVHN with 37\% cost reduction.

\begin{table}[!hbt]

	\begin{center}
	\renewcommand{\arraystretch}{1.2}
	\caption{Summary of Accuracy vs Cost reduction using SVHN, CIFAR10 datasets on PTEENet models.}
	\label{tab:costReduction}

	\begin{tabular}{|c|c|c|c|c|}

		\hline
		\hline
		Dataset & Network & \parbox{1.0cm}{\# Of\\ Branches} & \parbox{1.2cm}{Avg Val.\\ Accuracy\\ Reduction (\%)} & \parbox{1.2cm}{Avg Cost\\ Reduction (\%)} \\
		
		\hline
		SVHN & ResEEnet20 & 3 & 2 & 15 \\
		& ResEEnet32 & 5 & 0.5 & 15 \\
		& ResEEnet110 & 10 & 0 & 25 \\
		& DenseEEnet121 & 3 & 1 & 13 \\
		& VGGEEnet19 & 3 & 0.9 & 37 \\
		CIFAR10 & ResEEnet20 & 3 & 1.5 & 15 \\
		& ResEEnet32 & 5 & 1 & 15 \\
		& ResEEnet110 & 10 & 2.8 & 27 \\
		& DenseEEnet121 & 3 & 1.5 & 9 \\
		& VGGEEnet19 & 3 & 2 & 42 \\
	\hline
	\end{tabular}
	\end{center}
\end{table}  

\section{Summary}

In this work we proposed PTEEnet - a methodology for attaching and training early exit branches to pre-trained state-of-the-art deep neural networks. It has been shown that the output produced by the original network can be successfully used as labels for training the exits classifier and confidence heads, removing the need for the original labeled training data. Furthermore, we used a single confidence threshold parameter for controlling the accuracy vs cost tradeoff, allowing easy selection of an optimal point based on specific application requirements and constraints. Using several examples, we showed that a significant reduction in average computational cost can be achieved by selecting optimal confidence thresholds while sustaining only a small impact on the overall accuracy.

Although, the applicability of the approach is not limited to any specific task, the current work demonstrates the benefits of the method for image classification using several popular architectures of main network. The PTEEnet approach can be used alongside other neural network optimization techniques such as pruning and network compression methods that are usually performed on the main network.

Future work can explore more complex training methods. For instance, the branches head can be trained in an incremental manner with different fine-tuned confidence level thresholds for each exit. The threshold confidence level \emph{T} can be expanded to support a dedicated threshold level for each branch. This could result in further optimization of the accuracy-cost ratio and the control of data propagation through the branches in real-time.

\nocite{*}
\printbibliography[
heading=bibintoc,
title={References}
]
\end{document}